# Six-Degree-of-Freedom Motion Emulation for Data-Driven Modeling of Underwater Vehicles

Juliana Danesi Ruiz[1,2*], Michael Swafford[1,2], Austin Krebill[1], Rachel Vitali[1], and Casey Harwood[1,2]
([1]Department of Mechanical Engineering, The University of Iowa, [2]IIHR - Hydroscience and Engineering, The University of Iowa)

**ABSTRACT**

This article presents a collaborative research effort aimed at developing a novel six-degree-of-freedom (6-DOF) motion platform for the empirical characterization of hydrodynamic forces crucial for the control and stability of surface and subsurface vehicles. Traditional experimental methods, such as the Planar Motion Mechanism (PMM), are limited by the number of simultaneously articulated DOFs and are limited to single-frequency testing, making such systems impractical for resolving frequency-dependent added mass or damping matrices. The 6 DOF platform, termed a hexapod, overcomes these limitations by offering enhanced maneuverability and the ability to test broad-banded frequency spectra in multiple degrees of freedom in a single experiment.

The design, fabrication, control, and commissioning of a custom hexapod platform in the University of Iowa towing tank is described in detail. The hexapod's range of motion and dynamic position errors are carefully evaluated, demonstrating that the platform can support large payloads, move at high speeds, and maintain significantly sub-millimeter and sub-0.1 degree dynamic positioning errors, verified using a precision motion capture system. A machine vision system is described, using a hybrid fiducial and a hand-eye calibration to precisely estimate the hexapod's pose.

Experiments were conducted on a vertical truncated cylinder. A six-DOF load cell was used to measure the reaction forces when the cylinder was subjected to phase-optimized multisine motion profiles, from which frequency-dependent mass and damping matrices can be estimated. The results of the experiments were compared with results from a panel code simulation. After an exhaustive analysis of the hexapod system, it was found that a defective strain gauge in the load cell led to poor results compared to the validation data. However, the system identification methodology paired with the hexapod shows promise in the advancement of experimental hydrodynamic research in the ability to rapidly identify frequency-dependent force matrices.

**INTRODUCTION**

Naval engineers consistently face challenges in the design and characterization of submersible underwater vehicles. A key hurdle is the rapid and accurate identification of models, essential for analyzing fluid dynamic components that impact the performance and directional stability of these vehicles, including added mass and damping coefficients. The study of the added hydrodynamic components of fully and partially submerged bodies has been a pivotal topic of study for more than a century (Konstantinidis, 2013). It is critical in the equations of motion for underwater vehicles, as it requires the inclusion of additional hydrodynamic forces that are functions of both the object's geometry and the frequency of oscillation (Newman, 2018). Moreover, with the growing interest in autonomous systems, accurately determining these parameters becomes essential for designing precise controllers for autonomous operation.

Traditional methods used for the characterization of hydrodynamic derivatives are time-consuming and limiting. These include Planar Motion Mechanisms (PMMs), which are usually used for captive model testing in sway and yaw degrees of freedom (DOFs) (Ueno *et al.*, 2003), or the Vertical Planar Motion Mechanism (VPMM) used for heave and pitch motions (Lee *et al.*, 2011). While these methods offer valuable insights, they are constrained by their limited ranges of motion and require complex experimental setups relying on stepped sine perturbations of one or two DOFs at a time, which impedes rapid model identification. To address these limitations, The University of Iowa has designed and fabricated a state-of-the-art six-degree-of-freedom (6-DOF) motion platform to empirically characterize hydrodynamic parameters essential for robust control of autonomous underwater vehicles.

The Stewart platform, commonly known and hereafter referred to as a hexapod, was formally introduced in 1965 as an aircraft simulator (Stewart, 1965). The hexapod comprises two main bodies, a platform and a base, that are connected by six extensible legs, typically designed to be linear actuators (Jouini *et al.*, 2013). The actuators are attached to the platform and the base by two- or three-degree-of-freedom joints (Harib and Srinivasan, 2003). Given its parallel configuration, the hexapod has six degrees of freedom, capable of translating and rotating about all three dimensional axes, and providing a high load-to-weight ratio (Silva *et al.*, 2022). In the past 60 years, these characteristics have led the hexapod to be widely adopted in industries that require rigid motion platforms (Kazezkhan *et al.*, 2020). However, the parallel kinematics also make it challenging to ensure that adequate payload capacity and range of motion are maintain in all six-DOF.

Compared to PMMs, hexapods provide sophisticated motions, which Tu *et al.* (2019) showed to be effective for emulating the motions of free-running vehicles. By attaching a six-DOF force sensor between the hexapod's platform and the body, reaction forces in all DOFs can be measured and analyzed to properly populate a maneuvering model for such a vehicle. A reduced-order maneuvering model can be proposed as:

$$\mathbf{M_s}\vec{\ddot{X}} + \mathbf{C_s}\vec{\dot{X}} + \mathbf{K_s}\vec{X} = \vec{F_{EX}} + \vec{F_{FL}} \quad (1)$$

where $\mathbf{M_s}, \mathbf{C_s}, \mathbf{K_s}$ denote the mass, damping, and stiffness of the solid body, respectively. $\vec{X}$ is the displacement vector for each solid DOF. $\vec{F_{EX}}$ and $\vec{F_{FL}}$ represent vectors of externally applied forces and moments and fluid forces and moments, respectively. These fluid forces in equation 1 are typically unknown, and one of the principal challenges of fluid-structure interactions (and marine hydrodynamics) lies in identifying and modeling these fluid forces economically and accurately.

**Background: Hydrodynamic Loading**

As Newman (2018) outlines, unsteady motions produce hydrodynamic pressure distributions that are proportional in their magnitudes to the accelerations of a body and with spatial distributions that depend upon the shape and orientation of the body. This leads to the concept of added mass, which can be alternately derived under the assumptions of an ideal fluid or a viscous fluid at high Reynolds or Stokes numbers. Newman (2018) emphasizes the significance of the added mass coefficients "as the most important hydrodynamic characteristics of the body, except in the case of steady translation where ... the viscous drag is clearly more important." If the body is near a free surface or moving with low Reynolds numbers or Stokes numbers, a fluid force component can be derived that is proportional and phase-opposite to the body's velocities, termed the added damping. In the case of inviscid fluids, damping of single DOF only occurs near the free surface, where wave radiation dissipates kinetic energy from a body's motion (Newman, 2018). Hydrodynamic damping is significantly more challenging to quantify analytically than added mass, and is often studied empirically - though the plurality of damping models and the nonlinearity of real fluids pose obstacles to experimental estimation of damping. Finally, hydrodynamic lift or buoyancy (in the case of floating bodies) produces fluid forces that can be modeled as proportional to body displacements, dubbed the hydrodynamic stiffness.

This representation separates the fluid forces into terms that are – in part – proportional to the rigid body motions, leading to a lumped model of the form:

$$(\mathbf{M_s} + \mathbf{M_{FL}}(x,\dot{x}))\vec{\ddot{X}} + (\mathbf{C_s} + \mathbf{C_{FL}}(x,\dot{x}))\vec{\dot{X}}$$
$$+ (\mathbf{K_s} + \mathbf{K_{FL}}(x,\dot{x}))\vec{X} = \mathbf{B}\vec{\delta} + \vec{F_R} + \vec{F_{EX}}. \quad (2)$$

The matrices $\mathbf{M}$, $\mathbf{C}$, and $\mathbf{K}$ represent mass, damping, and stiffness, respectively, with the subscripts $s$ and $FL$ indicating respective contributions from the solid body and the fluid dynamics. The term $\vec{F_R}$, denotes the residual forces and moments resulting from steady and unsteady motion through water, such as hydrodynamic drag and unsteady vortex shedding. The term $\vec{F_{EX}}$ refers to the externally imposed forces and moments, and the matrix $\mathbf{B}$, a $(6 \times N)$ input matrix, maps the $N$ control actuations in $\vec{\delta}$ (e.g. rudder inputs) to the six forces and moments.

**Background: Frequency Domain System Identification**

A linearized model like equation 2 lends itself to linear system identification, wherein experiments are performed with inputs selected to perturb a measured response for each degree of freedom (Pintelon and Schoukens, 2012), enabling data-driven estimation of fluid hydrodynamic coefficients. Modern techniques in frequency domain system identification rely on employing broad banded signals over all DOFs using optimized excitation signals. The calculation of the response function of the system can be either parametric – based on a finite number of characteristics – or nonparametric – a system function at many points (Pintelon and Schoukens, 2012; Paraskevopoulos, 2002). While nonparametric models generally do not enhance one's knowledge of the system as it relates to physical laws, they are easier to design and implement (Pintelon and Schoukens, 2012). For the calculation of hydrodynamic coefficients absent of a specific maneuvering model, the nonparametric approach is sufficient for the engineer to evaluate the submerged vehicle dynamics.

The linear nature of this model representation enables the application of established linear system identification techniques to determine the unknown matrices $\mathbf{M_{FL}}, \mathbf{C_{FL}}, \mathbf{K_{FL}}$, and $\mathbf{B}$. By suspending an object from a six-DOF force and torque transducer and exciting the object in all six-DOF across a range of frequencies, it is possible to fully resolve the right-hand side of equation 2. Consequently, each of the unknown parameters can be isolated through a series of tests:

1. Swing table testing to determine **M<sub>S</sub>**.

2. Towing tank tests are conducted to record steady-state forces for each perturbed DOF, resulting in $\vec{F}_{FL}$ and one column of **K<sub>FL</sub>**.

3. Harmonic motions are induced using the hexapod. Analyzing the relationship between acceleration and measured forces uncovers **M<sub>FL</sub>**, **C<sub>FL</sub>**.

4. Harmonic motions for the actuators are similarly used to identify the **B**.

For the harmonic motions, so long as the system retains sufficient linearity – under small amplitude displacements and accelerations (Morelli, 2003; Pintelon and Schoukens, 2012) – it is possible to excite all DOFs of the system across a frequency band using different phase profiles during an experiment. By analyzing the modulus and phase of the transfer function between acceleration and force, one can fully identify the added mass and damping of the submerged body within a desired frequency band.

### Scope and Objectives

The aims of this paper are to (a) describe the design process for a hexapod with appropriate operating capabilities, (b) develop a system identification methodology to estimate frequency-depended system matrices, and (c) perform verification tests on a submersible object with known hydrodynamic properties. This work can significantly advance the efficacy of model-scale testing in both contrived and realistic environments. The hexapod is used to simulate free-running dynamics of the model for preliminary testing (Hassani *et al.*, 2015). The construction of the hexapod and the system identification effort that it enables are themselves of great relevance, as it establishes the capabilities to perform modeling of Navy submersibles, validate numerical models, and provide faster, more-precise estimates of maneuvering derivatives. Finally, as mentioned in previous sections, the hexapod allows new approaches to model characterization, rooted in dynamical system testing, that can dramatically improve experimental estimates of model dynamics with fewer trials required.

### METHODS

This section details the conceptualization, fabrication, programming, and governance of a six-DOF hexapod as well as the methods employed to ensure cohesive and optimal operation. In addition, it described the system identification equations and how the platform can be used to populate the fluid-dependent properties matrices.

### Hexapod Kinematic Modeling

A kinematic model is required to relate the platform position (the task space) to the actuator lengths (the joint space). Unlike serial manipulators, for which forward kinematics are relatively straightforward, fully parallel manipulators like the hexapod are much simpler to analyze through inverse kinematics (Harib and Srinivasan, 2003). The known variable for the inverse kinematics is the pose of the hexapod, or the three-dimensional Cartesian position and orientation of the hexapod's platform. For roll, pitch, and yaw Euler angles ($\theta, \phi, \Psi$), the following sequence of rotations is used: $\theta \rightarrow \phi \rightarrow \Psi$. The pose vector, $\vec{P}$, is thus defined as:

$$\vec{P} = \left[ x, y, z, \theta, \phi, \Psi \right]^\top \quad (3)$$

The hexapod joint space coordinate vector, $\vec{l}$, which consists of the lengths of all six actuators, is defined as:

$$\vec{l} = \left[ l_1, l_2, l_3, l_4, l_5, l_6 \right]^\top \quad (4)$$

The hexapod contains two main coordinate systems: 1) an inertial frame fixed to the base ($X_b, Y_b, Z_b$) and 2) a body frame fixed to the moving platform ($x_p, y_p, z_p$). At the home position, depicted in figure 1a, the base and platform frames (frame B and frame P, respectively) are aligned. During platform rotation, the rotation matrix, **R**, translates the platform frame to the base frame as:

$$\mathbf{R_{B/P}} = \begin{bmatrix} c\Psi c\phi & c\Psi s\theta s\phi - sin\Psi c\theta & c\theta c\Psi s\phi + s\Psi s\theta \\ s\Psi c\phi & s\Psi s\theta s\phi + c\Psi c\theta & s\Psi s\phi c\theta - c\Psi s\theta \\ -s\phi & c\phi s\theta & c\theta c\phi \end{bmatrix} \quad (5)$$

where $s$ denotes the sine function and $c$ denotes cosine.

For the hexapod inverse kinematics, the closed-loop vector approach (Silva *et al.*, 2022; Pedrammehr *et al.*, 2011; Tsai, 2000; Kazezkhan *et al.*, 2020) and the hexapod rotation matrix, equation 5, are used to determine the link lengths. Specifically, the following equation is used:

$$\vec{l_i} = (\vec{p_{CM}} + \mathbf{R_{B/P}} * \vec{p_i}) - \vec{b_i} \quad (6)$$

where $\vec{p_i}$ is the position of the actuator platform attachment points ($i = 1, 2, 3...6$) to the origin of the platform, $\vec{b_i}$ is the position of the actuator base attachment points to the origin of the base minus the height to the axis of the U-joint yoke attachment, and $\vec{p_{CM}}$ is the position of the origin of the platform frame from the origin of the base frame. This way, each actuator length can be calculated based from the platform pose as the Euclidean norm of the link vector $||\vec{l_i}||$. By using this method to compute the inverse

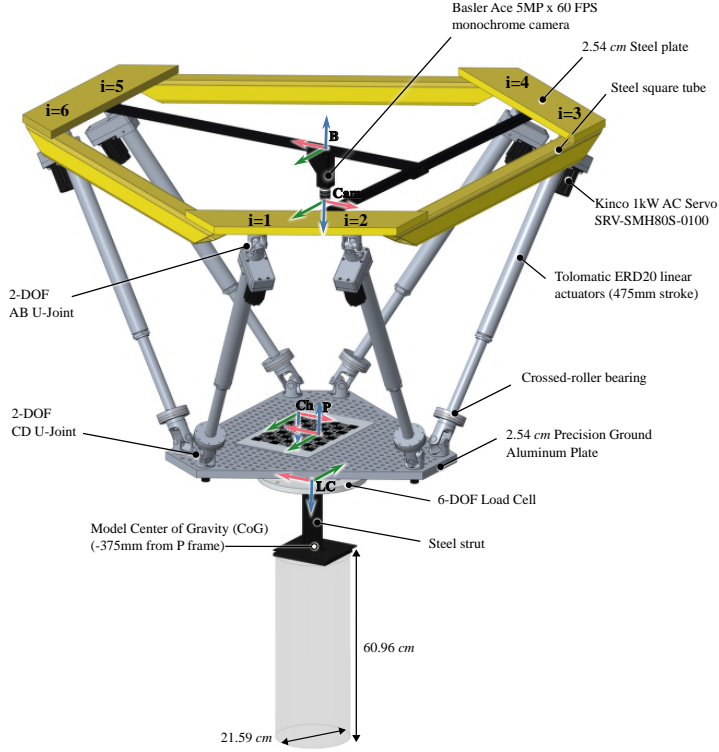
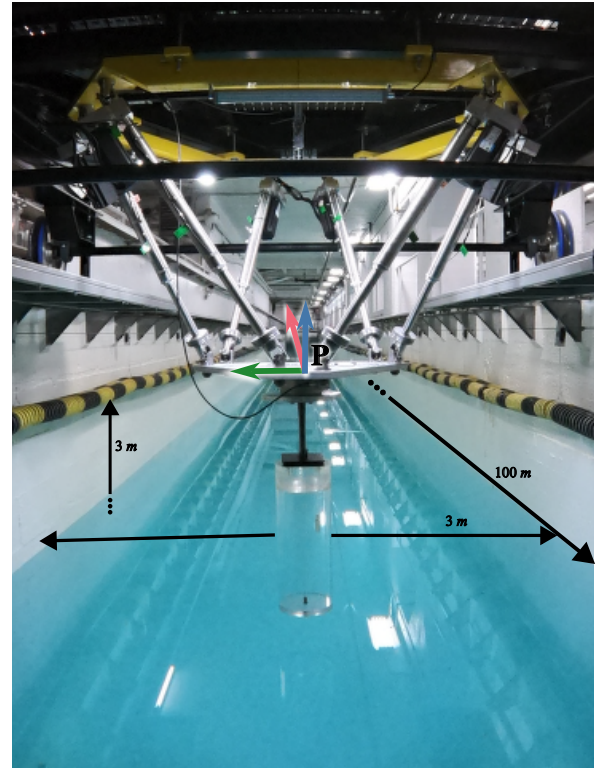

**(a)** 3D model of the hexapod's experimental configuration

**(b)** IIHR Towing Tank facility

**Figure 1:** (a) The experimental setup is illustrated with callouts to major components and instruments. The system multiple stationary and moving coordinate frames are also illustrated, with +X (red), +Y (green), and +Z (blue). (b) The towing tank facility, annotated with dimensions of 3 m x 3m x100 m (w x d x l) and orientation.

kinematics, it allows us to modify the $\vec{p}_{CM}$ of the platform in the case of a object being attached to the platform to prescribe motions with respect to an alternate coordinate system (e.g. the center of mass of an attached model).

**Dynamic Modeling**

To assist in the selection of hexapod components, the dynamics of the manipulator were also computed to ensure the resulting actuator loads were within manufacturer limits. The inverse dynamics model adopts a quasi-static assumption (massless actuators), with loads that include the weight and inertia of the model and hexapod platform, as well as any externally applied (e.g. hydrodynamic) forces. A linear mapping of axial actuator forces to forces and moments about an arbitrary point in space can be written through the system of equations (Innovations, 2023),

$$\vec{F}_{ext} = \sum_{i=1}^{6} \hat{l}_i F_i, \qquad \vec{M}_{ext} = \sum_{i=1}^{6} \vec{r}_{p,i} \times \hat{l}_i F_i, \quad (7)$$

where $\vec{F}_{ext}$ and $\vec{M}_{ext}$ are the forces and moments applied at some known point, relative to the hexapod platform, $F_i$ and $\hat{l}_i$ are the force (positive tension) and unit vector of actuator $i$, and $\vec{r}_{p,i}$ is the vector describing the distance from actuator attachment point $i$ to the point of external load application. The equation can be compactly stated as,

$$\mathsf{T_L} \vec{F} = \left\{ \vec{F}, \vec{M} \right\}_{ext}^{T}, \quad (8)$$

where $\mathsf{T_L}$ is the transformation from actuator axial forces to the forces and moments applied at a point, $\vec{F}_{ext}$. Inversion of this equation permits the forces on each actuator to be calculated as a function of the applied loads and the instantaneous pose of the hexapod.

**Joint Kinematics**

The hexapod described in this paper was built using two-DOF universal joints on each end of all six linear electric actuators. Crossed-roller bearings are installed between the actuator rod-ends and the platform universal joints to permit rotation about each actuator's axis, as shown in figure 1a. This design allows for three DOF at the platform joint, effectively acting as a spherical joint. This configuration, known as "6-UPU" (Kazezkhan *et al.*, 2020), provides a higher payload capacity than configurations using true spherical joints. Unfortunately, during preliminary range-of-motion testing, a pair of U-joint yokes collided, producing a large axial torque that damaged two actuators. To prevent future collisions and better characterize the hexapod's range of motion a kinematic model was introduced for the U-joints to check for collisions. Figures 2a and 2b illustrate the vector notation of the AB joint (U-joint attached to the base) and the CD joint (U-joint attached to platform).

For the AB joint shown in figure 2a, the unit vectors $\hat{u}_i$ and $\hat{l}_i$ are known quantities (i.e., they are part of the known geometry of the hexapod). Specifically, the first axis of rotation for the AB U-joint is a fixed unit vector and can be found from the hexapod geometry. The first yoke is fixed at a 20 degree angle to the platform edge. The $\hat{l}_i$ vector is the direction unit vector of the actuator, found by normalizing the actuator link vector from equation 6. Then, the $\hat{v}_i$ vector represents the second revolute joint axis of the AB U-joint and due to the U-joint cross format, it is in a plane normal to $\hat{u}_i$ and to $\hat{l}_i$ (Pedrammehr *et al.*, 2011). Finally, the $\hat{c}_i$ vector is in a plane perpendicular to the $\hat{u}_i$ and $\hat{v}_i$ vectors.

The joint kinematics described were used to calculate the rotational angles for the two U-joint, AB and CD, on each actuator. MATLAB mesh collision detection was used to check for collisions and compute the minimum distance between the U-joint yokes in each pair for rotation angles between 0 and 90 degrees about each axis in increments of 0.1 degree. During operation, U-joint angles are computed as part of the hexapod inverse kinematic model, which are used to interpolate the clearance before U-joint yokes. A contour plot depicting the yoke clearance for joint AB is shown in figure 3.

**Physical Design and Electronics**

The kinematic and dynamic models of the hexapod were used to drive component selection by simulating realistic loading and motion scenarios to evaluate the suitability of actuators with respect to thrust, speed, and acceleration capabilities. An overview of the final design components is illustrated in figure 1a. Tolomatic ERD20 linear actuators with a stroke length of 475 mm were selected as the hexapod legs, connecting the 0.8-meter diameter platform and the 1.6-meter diameter base. The actuators are positioned in a circumferential pattern by spatially distancing alternating actuators by 100 degrees and 20 degrees. This arrangement adheres to the CSSP (Circle Single Stewart Platform) criterion, which is crucial to ensure design symmetry (Silva *et al.*, 2022).

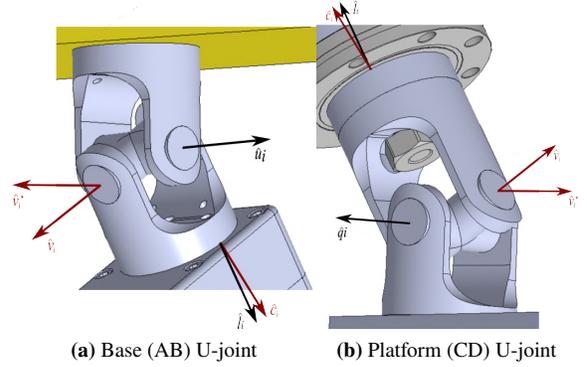

**(a)** Base (AB) U-joint  **(b)** Platform (CD) U-joint

**Figure 2:** (a) AB Base Universal Joint vector notation (b) CD Platform Universal joint vector notation. Black vectors denote quantities that are known from the geometry or the commanded pose of the hexapod. Red vectors denote quantities that are calculated for the joint kinematics.

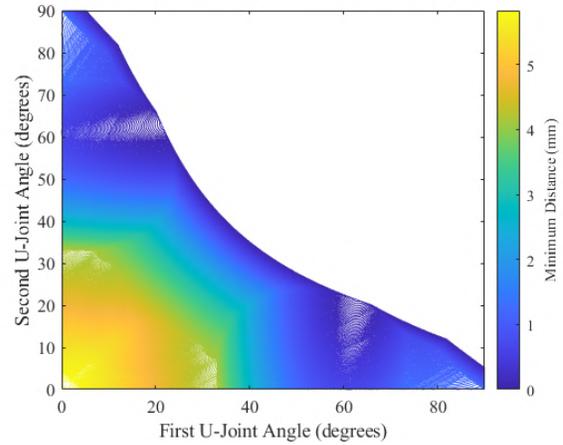

**Figure 3:** A comprehensive contour plot of the lookup table to avoid collisions, delineating the minimal distance as fine as 0.1 millimeters

Each of the six actuators is rated for 2.2 kN of thrust, with a maximum speed of 0.82 m/s and an acceleration of 8.9 m/$s^2$. The actuators are paired with Kinco SRV-SMH80S-0100 motors with passive brakes and driven by Kinco-SRV-FD332S-LA-000 servo drives, rated for 1kW output and a maximum speed of 5500 RPM. Each drive is fitted with a 400-watt brake/chopping resistor to assist with dissipation of large inertial loads. The motors include 10,000 count-per-revolution incremental encoders, resulting in an axial motion resolution of 25 microns for each actuator. Each actuator is fitted with three hall-effect switches: positive limit, negative limit, and homing.

The base of the hexapod is a steel weldment structure, which was faced on a large gantry mill to ensure that all attachment points are within approximately 25$\mu$m of plane. The platform is cut from a solid piece of 2.54 cm (1 inch) thick 6061 aluminum, which is precision ground on both sides. The platform was drilled and tapped with 3/8-inch holes in a 2.54 cm (1 inch) grid aligned with the hexapod coordinate axes. The 24 U-joints yokes were milled from 2.5-inch 4140 steel rod, using a five-axis mill, with Yamaha ATV-202 roller-bearing universal joints.

The six drives and their respective limit switches are connected to a Galil DMC-4080 motion controller via custom-fabricated junction PCBs that facilitate routing of encoder, command, and limit-switch signals. The power to the hexapod is provided by a dedicated 100-amp, 110V subpanel, with each motor on a dedicated, filtered 15-amp circuit. The motor drives and Galil motion controller are powered via individual 24V DC power supplies that reside in another dedicated 15-amp circuit. Line level AC circuits, DC circuits, and control circuitry are segregated into separate shielded enclosures to maintain signal integrity.

**Control and Programming**

The Galil motion controller is used as the PID position controller for all six actuators, with gains adjusted to minimize actuator position errors during motion under load. This leaves the Kinco servo drives to act as open-loop torque amplifiers. The communication between user and the Galil uses a custom MATLAB graphical user interface (GUI), named HUGO (Hexapod User Guidance and Operation). The high-level overview of the HUGO workflow is illustrated in figure 4. It enables the user to specify a movement command in Cartesian pose space and outputs readable joint space data to the Galil controller. This is achievable as shown in the 'Simulate Motions' section by implementing the hexapod kinematics to compute the actuator lengths for each pose, the U-joint kinematics to calculate the minimum separation distance between the yokes, and the hexapod dynamics to estimate the axial load for each actuator. The results of the 'Simulation Motions' are then subjected to various safety checks to ensure actuators, motors, drives, and U-joint angles are all within rated/safe limits, as demonstrated in the 'Check Constraints' section. The last section, named 'Execute Motions' summarizes how actuator position values are then used in a PID loop to actuate the six legs. Thus, the MATLAB GUI effectively serves as a digital twin of the hexapod, providing a user-friendly interface to control the hexapod's movements.

**Camera-based tracking**

A sophisticated camera-based pose tracking methodology was implemented to ensure alignment between platform tracked and commanded poses during testing. A Basler Ace 5MP x 60 FPS monochrome camera was paired with a Ricoh F6mm Lens to achieve the desired field of view.

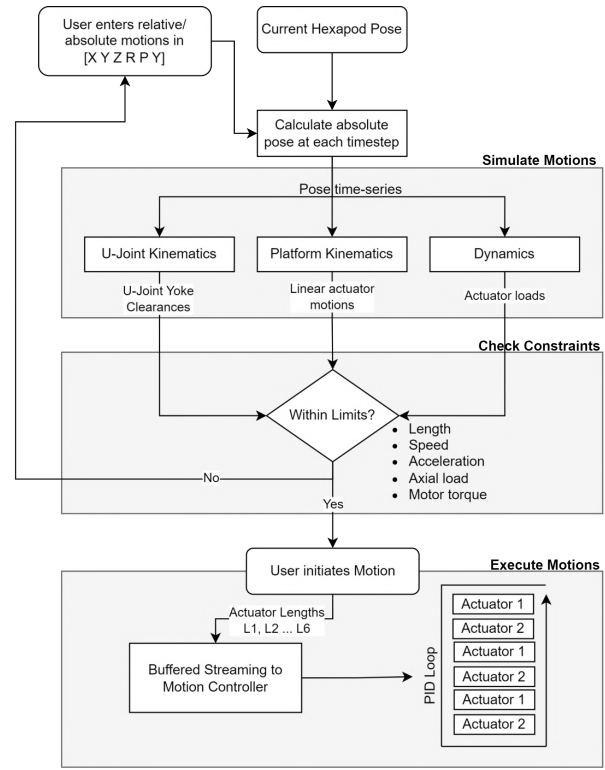

**Figure 4:** Flow-chart of the hexapod motion planning process. The input consists of the hexapod Cartesian pose and the output is the hexapod joint space.

For accurate tracking of the platform position, a ChArUco board was installed on the camera-facing surface of the platform, as depicted in figure 1. The ChArUco board amalgamates a chessboard pattern with ArUco markers (ChArUco = Chessboard + ArUco), integrating the benefits of both approaches. This combination exploits the position accuracy benefits from the Chessboard adjacent black squares pattern (Lin *et al.*, 2022), and the robustness and reliable tracking of ArUco due to its unique encoded marker. Furthermore, the target can be properly detected even with partial occlusion (i.e., when the ChArUco board attached to the platform travels partially outside of the camera's field of vision) and suboptimal conditions (e.g., poor lighting) (Enebuse *et al.*, 2021).

The camera calibration and subsequent platform tracking were conducted using the ChArUco board. The board measures 254x381 mm, covering a significant portion of the platform, with square length of 30mm and marker (ArUcO) length of 22mm, arranged in 9 columns and 7 rows. This configuration ensured that the board would be able to be properly captured by the camera and detected at any platform position. The calibration process utilized the extensive OpenCV library (Bradski, 2000) to determine intrinsic camera parameters (e.g., the camera matrix and the radial and tangential distortion coefficients), to accurately interpret the image captures by the camera sensor, and to correct any distortions (Jütte *et al.*, 2023). The calibration's efficacy is assessed by computing the reprojection error, a key metric in quantifying the accuracy in which the 2D image plane is mapped into the 3D world (de Medeiros Esper *et al.*, 2022).

The tracking methodology is summarized by figure 5, which delineates the two distinct coordinate frame worlds: Hexapod World and Camera World. As depicted, the homogeneous transformation matrix of the hexapod platform relative to its base is given by **T$_{P/B}$(t)**, encapsulating both the rotation and translation of the platform at the $t^{th}$ time step. Next, the transformation matrix, **T$_{Ch/Cam}$(t)**, represents the pose of the ChArUco board with respect to the camera frame at the same $t^{th}$ time step, estimated using the OpenCV library ChArUco board pose estimation function. The primary objective of the tracking is to estimate **T$_{P/B}$(t)** based on **T$_{Ch/Cam}$(t)**. However, these frames are in two separate coordinate worlds, precluding a direct relationship to be established.

To bridge this gap between coordinate worlds, a hand-eye calibration is performed. The hand-eye calibration problem can be concisely formulated as **AX = XB**, in accordance with Tsai's seminal work (Tsai *et al.*, 1989). The eye-to-hand calibration involves a stationary camera (e.g., fixed to the hexapod base) and a moving target (e.g., the ChArUco board), affixed to the robot moving gripper (e.g., the hexapod platform). It aims to compute the transformation matrix between the two fixed coordinate world frames, the camera and the hexapod base, denoted as **T$_{B/Cam}$**. The data used for this calibration is systematically collected across multiple motion profiles covering all six-DOF, then the transformation matrix, **T$_{B/Cam}$**, can be estimated using the OpenCV library and the hand-eye method proposed by (Daniilidis, 1999). Unlike Tsai's approach, this method utilizes dual quaternions for an improved calibration that simultaneously compute rotation and translation.

After a transformation between the two worlds is formed through the hand-eye calibration method by estimating **T$_{B/Cam}$**, the other fixed frame **T$_{P/Ch}$** can be computed. The ChAruco board is attached to the hexapod platform, fixing the relationship between the ChArUco frame and the platform frame. This transformation can be computed by taking advantage of transformation matrices multiplication and minimizing noise. The final estimated transformation from the hexapod base to platform position can then be described by equation 9:

$$\mathbf{T_{P/B}}(t) = \mathbf{T_{P/Ch}} * \mathbf{T_{Ch/Cam}}(t) * \mathbf{T_{Cam/B}} \quad (9)$$

where **T$_{P/Ch}$** and **T$_{Cam/B}$** are fixed frames estimated using the hand-eye calibration and matrix multiplication.

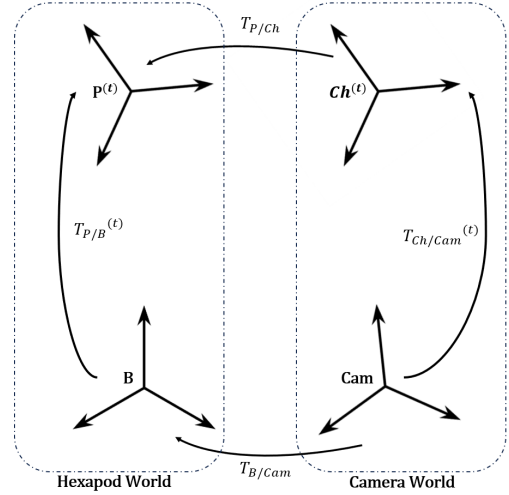

**Figure 5:** Relationship between the coordinate frames of the 'Hexapod World' and the 'Camera World' to perform the Hand Eye Calibration. In Hexapod World, 'B' is the fixed hexapod base, '**P(t)**' is the moving platform. In Camera world 'Cam' is the fixed camera frame, and '**Ch(t)** is the moving ChArUco frame.

### System Identification

To achieve an accurate estimate of the added mass matrix **M$_{FL}$** and added damping matrix **C$_{FL}$** for a submerged body, an orthogonal multisine excitation signal with optimized crest factor was designed for the hexapod, and the robust local polynomial method was used to estimate the frequency response matrix (FRM), which is a tensor made up of individual frequency response functions (FRFs). The following paragraphs detail the experimental procedures and data processing techniques used to identify the added mass and damping for a submerged body. The methodology for system identification adheres closely to the best practices and techniques outlined in Pintelon and Schoukens (2012), which provides a comprehensive treatment of the subject for the interested reader. As stated previously, a nonparametric approach to estimating the FRM is sufficient to recover the added mass and damping for a submerged body. Therefore, according to Pintelon and Schoukens (2012) the optimized excitation signal in the nonparametric case should be one that gives the most accurate estimate in a set amount of time, without exceeding a specified maximum signal limit:

$$\min(\max_{k \epsilon \mathbb{F}} \sigma_G^2(k)) \quad with \quad \max_t |u(t)| \leq u_{max} \quad (10)$$

Where $\sigma_G^2(k)$ is the uncertainty of the FRF at $\omega_k$. Weak signals lead to uneven distribution of signal energy across the desired frequency spectrum (Pintelon and Schoukens, 2012). The goal is to maintain constant uncertainty at all frequencies, which requires adjusting the power distribution of the signals based on the level of background noise

(Pintelon and Schoukens, 2012). In the literature, this leads to two important metrics: the crest factor and the time factor (Pintelon and Schoukens, 2012). The crest factor defines the ratio of the signal's peaks to its average. Signals with a lower crest factor inject more power into the system, improving the persistence of the excitation signal and leading to more knowledge of the system (Pintelon and Schoukens, 2012). The time factor defines the time measurement required per frequency point to maintain some desired minimum signal-to-noise ratio (Pintelon and Schoukens, 2012).

The excitation signal designed for use in this work was the full random orthogonal multisine, including optimization of the crest factors of the excited harmonics with respect to the signal (displacement) and its second derivative (acceleration). Optimization of the crest factor is uses the $l_{2p}(\phi)$ norm from Guillaume *et al.* (1991); Pintelon and Schoukens (2012):

$$\left\| \frac{u(t,\phi)}{u_{rms}}, \frac{\ddot{u}(t,\phi)}{\ddot{u}_{rms}} \right\|_{2p} = \left[ \frac{1}{T_0} \int_0^{T_0} \left( \frac{u^{2p}}{u_{rms}^{2p}} + \frac{\ddot{u}^{2p}}{\ddot{u}_{rms}^{2p}} \right) dt \right]^{\frac{1}{2p}} \tag{11}$$

where $u(t,\phi)$ and $\ddot{u}(t,\phi)$ are the displacement and the acceleration respectively, $\phi^T = [\phi_1, \phi_2...\phi_F]$ are the optimum phases, $T_0$ is the sampling period, and $p$ is a constant that determines the order of the norm. The $l_{2p}$ norm in equation 11 seeks to adjust the phases of displacement and acceleration of the excitation signal such that the crest factor of both approaches its optimum at $\sqrt{2}$. Both displacement and acceleration of the signal are considered to avoid excessive forces in the output measurement while keeping the stroke of the hexapod actuators small to maintain the linearity of the system.

The full random orthogonal phase-optimized multisine signal offers several advantages over other broad-band signals for system identification: it is periodic (no spectral leakage), has a flat power spectrum (delivering constant energy across the desired frequency band), and possesses a low crest factor and time factor due to the optimization. The algorithms designed to calculate the phase-optimized multisine for the multiple-input-multiple-output (MIMO) multisine used in the hexapod testing were written in MATLAB using an adaption of the calculate random phase orthogonal multisine written by Pintelon and Schoukens (2012) along with the optimization tool box. The algorithm is first initialized with a set of unique random phase profiles for each DOF and each experiment of size *Number of Inputs × Number of Inputs × Number of excited harmonics*. For the six-DOF hexapod system, this results in six experiments with six input excitations. The algorithm then generates a multisine for each experiment and degree of freedom from which the crest factor for each DOFs displacement and acceleration are calculated. The difference between the sum of these crest factors and the optimum crest factor ($\sqrt{(2)} * n_u^2$) is fed to the optimizer function (e.g. MATLAB's *fmincon*). The output of the optimization is the set of optimal phases, which can be fed into the multisine generator algorithm itself to generate the optimal test signal. This is done for a minimum of $M = 2$ realizations of the phases to generate $M$ unique realizations of the measured FRM that can be averaged for the best estimate of the system FRM. Utilizing multiple realizations improves noise rejection on the transfer function estimate – "the frequency resolution improves from $f_s/N$ to $M \cdot f_s/N$, and the results are averaged over the $M$ realizations" (Pintelon and Schoukens, 2012). This method ensures orthogonality among all DOFs tested in each MIMO experiment, where each experiment uses a unique, mutually orthogonal combination of motions. Thus, the influence of motion in each DOF on forces and moments in each other DOF remain separable, ensuring that any off-diagonal terms in the hydrodynamic mass and damping matrices are not conflated with terms along the diagonal.

These algorithms were packaged into a dedicated user interface called the Hexapod Multisine Generator (HMG) application, allowing for the creation of phase-optimized multisines for use with the hexapod. Because the robust local polynomial method necessitates a minimum of at least two periods of the excitation signal to estimate the non-periodic noise covariance, the generated excitation signal is replicated over the desired number of periods. These excitation signals are then transferred to the HUGO application for simulation and transmission to the hexapod. In this work, experiments were executed consecutively, with no time allocated between. Data from each experiment include the commanded displacements (input) and the forces and moments (output).

The entire set of data collected is processed using the robust local polynomial method (Pintelon and Schoukens, 2012). This method assumes that the transfer function matrix $\mathbf{G}(\mathbf{\Omega})$, along with the system and transient noise terms, are smooth functions of frequency whose quantities can be approximated in the local frequency band $[k - n, k + n]$ by a polynomial. The result is an estimate of the FRM $\mathbf{\hat{G}}(\mathbf{\Omega_k})$, and the noise covariance matrix $\mathbf{\hat{C}_v(k)}$ at frequency $k$ (Pintelon and Schoukens, 2012). This procedure is repeated for every frequency in the Discrete Fourier Transform (DFT), yielding an estimate of the FRM and noise covariance matrix at each frequency. These matrices are then concatenated to form a single estimate of the FRM and noise covariance for the experiment. by averaging all experiments, an accurate estimate of the FRM and noise covariance of the system is obtained.

From the estimated FRM, the added mass and

damping can be determined using the following:

$$\frac{\vec{a}}{\vec{f}} = \hat{\mathbf{G}}(\Omega) = \frac{1}{\mathbf{M}' - \frac{i}{\omega}\mathbf{C}_{\mathbf{FL}}} \quad (12)$$

where $\vec{a}$ and $\vec{f}$ represent the complex valued commanded acceleration and force and torque vectors, respectively. Here, $\mathbf{M}'$ is the total mass (solid mass + added mass), and $\omega$ denotes a single test frequency. By solving for the total mass and the added damping the following equations are obtained:

$$\mathbf{M}' = Real(\frac{1}{\hat{\mathbf{G}}(\Omega)}) \quad (13)$$

$$\mathbf{C}_{\mathbf{FL}} = imag(\frac{1}{\hat{\mathbf{G}}(\Omega)}) \cdot -\omega \quad (14)$$

Equations 13 and 14 are solved at each frequency in the FRM. To get the added mass the solid mass is subtracted from the solution for the total mass across the frequency band of interest. The derived values can then be applied to any dynamic model of the system, such as a coefficient-based model like the Gertler and Hagen equations (Gertler and Hagen, 1967), a look-up-table model, a lumped mass model, a state-space model, or any other mathematical dynamics model. Utilizing one or a combination of these models, a controller for the UUV can then be constructed. Thus the hexapod system allows for the FRM of a submerged body to be identified across $n_u$ experiments. Tests using monotone sinusoidal motions that sequentially test each DOF and each relevant frequency require experiments that grow in number linearly with the number of frequencies ($O(f_k)$). In contrast, the hexapod system combined with multisine excitation reduces to a constant number of experiments ($O(1)$).

**Table 1:** Summary of hexapod verification testing with vertical truncated cylinder. The state refers to whether the cylinder was tested unsubmerged (dry) or submerged (wet).

| Type | State | Freq. (Hz) | Amp. (mm) | Total Trials |
|---|---|---|---|---|
| Single Frequency Surge | Dry | 1 | 5 | 1 |
| SISO Multisine Surge | Dry | 0.4 - 2.35 | 10 | 2 |
| SISO Multisine Surge & Sway | Wet | 0.4 - 2.35 | 10 | 2 |
| MIMO Multisine | Dry | 0.4 - 2.35 | 5 | 12 |
| MIMO Multisine | Wet | 0.4 - 2.35 | 5 | 12 |

**Testing Facility and Experimental Procedure**

The hexapod was installed on an unpowered carriage at the University of Iowa IIHR towing tank as shown in figure 1b. The IIHR towing tank measures 3m x 3m x 100 m (width x depth x length) with a drive carriage capable of towing models up to 3 m/s. The hexapod carriage is pushed by the main drive carriage via two removable linkages. As depicted in figures 1a and 1b an ATI Omega 160 six-DOF load cell was installed on underside the hexapod platform to measure reaction forces and torques during prescribed motions. Data from the 6-DOF load cell was logged using a National Instruments cDAQ-9179 data acquisition (DAQ) chassis with two NI 9239 24-bit analog input modules. A synchronization trigger from the Galil motion controller to a NI-9402 counter module was used to time-align load cell measurements with the motions of the platform.

A vertical truncated cylinder was selected as a test geometry, shown in figure 1a. The cylinder is made of acrylic tubing that is capped at both ends and measures 21.59 cm in diameter and 60.96 cm in length. The cylinder is attached to the six-DOF load cell via a steel strut with an elliptical cross section measuring 25.4 cm in length. The total "sprung" mass of the cylinder and mounting hardware was measured at 16.67 kg and its moment of inertia in roll and pitch was measured using a swing table. The center of gravity was determined to be coincident with the cylinder top, 37.5 cm below the hexapod platform origin.

A total of five verification tests were conducted to commission the hexapod in the IIHR towing tank, which spanned single-frequency and multi-frequency tests in single and multiple degrees of freedom with the cylinder hanging in air and submerged in water. Table 1 lists these conditions. Monotone surge tests were used to verify that motions and forces were synchronized and resulted in the expected reaction force and torques for the given acceleration. Single-input-single-output (SISO) multisine dry tests were used to verify that the phase-optimized multisine accurately captured the known mass with the hexapod system over a desired band of frequencies. SISO multisine wet tests were used to verify that the hexapod and system identification procedure was able to recover the added mass and damping of the submerged cylinder system. MIMO multisine tests were likewise used to verify that the full mass and damping matrices were recoverable using the system identification methodology. Results from the dry tests were compared to the measured mass and moments of inertia from swing table tests. Results from the wet tests were compared to panel code simulations for a cylinder of the same geometry and depth.

The testing procedures for each type of tests were similar in design, but varied slightly in detail. For the single frequency tests, the HUGO application was used to directly design the input signal, whereas the multisine signals were designed and exported from the HMG application. The number and length of experiments was determined by the number of DOFs being excited and the desired frequency resolution. For the single frequency tests this resulted in collection times of 15 seconds, for the SISO multisine tests two experiments of 202 seconds each were performed, and for the MIMO multisine tests twelve experiments of 202 seconds each were performed. The following framework summarizes the testing procedure for each experiment with the hexapod at the home position and motions specified about the sprung masses center of gravity:

1. Generate excitation signal
2. Simulate dynamics of the signal in HUGO
3. Verify signal is within safety limits
4. Tare the load cell
5. Arm the load cell for recording
6. Execute Motion

Data recorded for each experiment included the hexapod designed motion profile, the measured forces and torques, and for some experiments the camera tracking data.

## RESULTS & DISCUSSION

This section presents the results for the hexapod performance characterization and verification trials in the towing tank. Presented first is the performance characterization of the hexapod detailing its reachable space, accuracy of positioning during dynamic testing, and analysis of the camera-based tracking and pose estimation. Results from the verification trials on the vertical truncated cylinder follow.

### Hexapod Performance Characterization

To evaluate the construction and programming of the hexapod, a series of comprehensive tests were conducted, focusing on its range of motion and dynamic positioning capabilities. The hexapod was engineered with precision by orchestrating six linear Kinco actuators with a Galil controller and kinematics equations. The actuation is managed through the HUGO application, which enables the user to input the desired platform pose – specified by coordinates $[x, y, z]$ and orientation angles $[\theta, \psi, \Phi]$ – or a sequence of poses. The GUI then applies the inverse kinematics equations to calculate the necessary actuator length to achieve the desired coordinate position and angle.

To assess the hexapod's reachable space, each degree of freedom was separately tested by incrementally maneuvering the robot. The empirical findings described in Table 2 reveal the extensive range of motion of the hexapod for all degrees of freedom. The range is subject to two constraints:(1) the maximum and minimum actuator lengths, and (2) angular limits of the U-joints. Translations encounter predominantly restrictions due to the actuator limits, while the rotations are mostly limited by the U-joint, as evident in Table 2.

**Table 2:** Tabulated range of motion for all six degrees of freedom

| | Translation (meter) | | |
|---|---|---|---|
| | X | Y | Z |
| **Max** | 0.35* | 0.33* | 0.28* |
| **Min** | −0.34* | −0.33* | −0.25* |

| | Rotations (degrees) | | |
|---|---|---|---|
| | Rx | Ry | Rz |
| **Max** | 33† | 22† | 48* |
| **Min** | −33† | −34† | −48* |

*Limited by actuator extension (limit switch).*
†*Limited by U-joint angles.*

The hexapod's accuracy during dynamic motions was examined using an optical motion capture (MoCap) system. The system consists of six OptiTrack Prime 17 cameras, arranged in a close formation around the hexapod , providing marker tracking with an estimated residual of 0.3 mm. A total of eight markers were affixed to the hexapod's platform to constitute a rigid body with an origin that coinciding with the hexapod's tool frame. Three additional markers were positioned on the hexapod's fixed base frame to define the inertial base frame. A series of motion sequences were performed to evaluate discrepancies between the commanded poses and the recorded motions observed using the MoCap system. The position error between commanded motion and MoCap data were remarkably low. Across motion types, frequencies, and platform payloads, the root-mean-square translation errors were consistently and significantly less than one millimeter and angular errors were less than 0.1 degree, validating that the kinematics are accurately computed.

Due to the open-loop control of the hexapod, a camera-based tracking algorithm is used to monitor the true pose for position errors. Tracking consists of using a ChArUco board to precisely detect hexapod poses and the results of the hand-eye calibration to estimate the platform pose in the hexapod base frame. Figure 6 illustrates the MoCap and the camera-based tracking results for a coordinated translation in X,Y,Z degrees of freedom. Comparing the camera-based tracking outcomes with the commanded hexapod movements, the RMSE for translation was found to be less than 1 mm, and the rotation errors remained less than 1 degree. The small rotations are thought to be a a result of an imperfect lens distortion model, which may be correctable with an improved lens calibration and/or a lower-distortion lens.

### Verification Trials

As described in the Testing Facility and Experimental Procedure section, five different verification trials were conducted to commission the hexapod in the IIHR towing tank. The first of these tests was the single-frequency tests where the dry cylinder was moved sinusoidally in surge with a frequency of 1 Hz and an amplitude of 5 cm. The designed acceleration amplitude from these tests was 1.974 $m/s^2$, leading to an expected reaction force of $F_x = 32.77$ N and an expected moment in pitch about the load cell origin of $M_y = 8.31$ Nm. Figure 7 shows the measured reaction force and torque, the solid lines represent the raw measurements, the black-dashed lines show $F_x$ and $M_y$ force and torque after a filter is applied, the expected force and torque are shown as the purple horizontal lines. As can be observed, the amplitudes of the filtered force measurements closely match the predicted values.

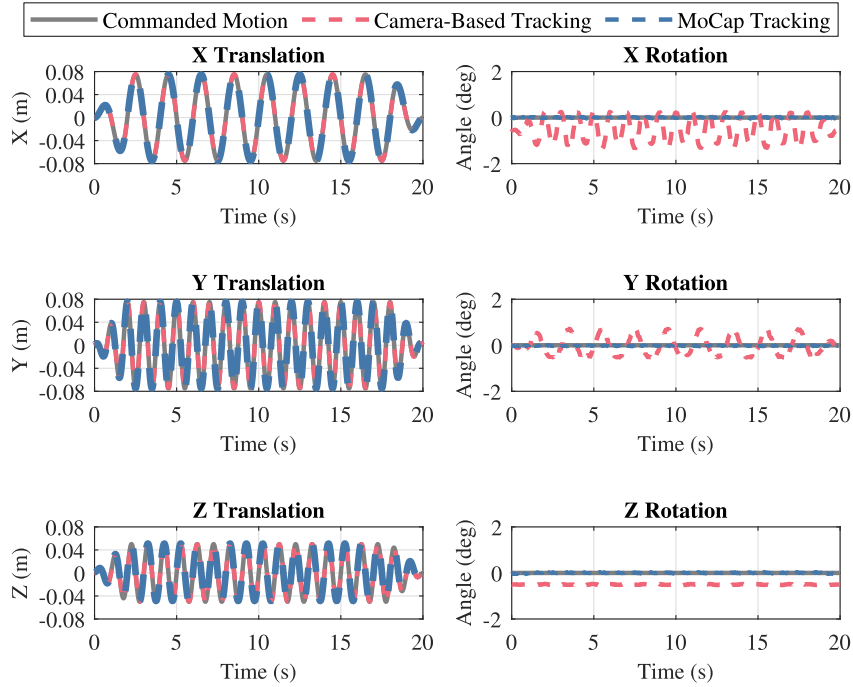

**Figure 6:** Example of MoCap and camera-based tracking for a coordinated translation in X,Y, and Z. RMS errors in translation are very small, and errors in rotation are acceptably small.

Following the monotone frequency test the SISO multisine surge tests were conducted in dry conditions, with the goal to accurately recover the mass of the cylinder over the frequency band $[0.4, 2.35]$ Hz. Two realizations of ten periods of with $T_0$ equal to 20 seconds were performed, resulting in a frequency resolution of 0.05 Hz. The total testing time to measure the FRF at 40 frequency points was 6 minutes and 44 seconds of testing time, and closer to 10 minutes when including time spent loading and simulating the excitation signal in the HUGO application. Figure 8 depicts an example of the multisine signals used. It shows a very compact signal with ten total periods. The lower two plots also show the ramp-in and ramp-out segments of one-second duration, represented by the dashed lines. These ramp segments are needed to smoothly transition between a resting state and the multisine motions, and are prescribed as polynomials with boundary conditions to ensure C1 continuity between the ramp segment and the beginning of the multisine. The ramps are discarded during the system identification process.

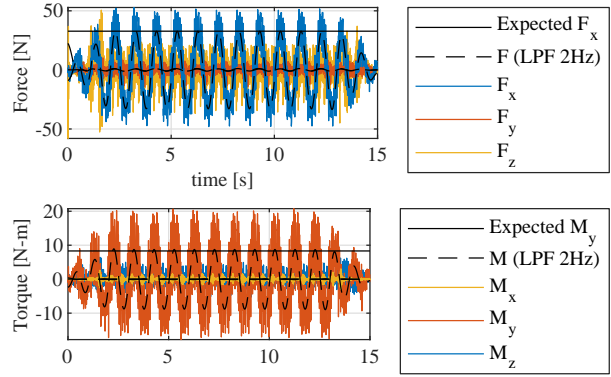

**Figure 7:** Force and torque measurements from surge monotone frequency tests of amplitude 5 cm and frequency 1 Hz. The solid black lines show the expected values of 32.77 N and 8.31 Nm for $F_x$ and $M_y$ respectively. The black-dashed line is a low-pass-filtered measurement of $F_x$ and $M_y$ with cutoff frequency 2 Hz. The result shows that the motion produced the expected measured reaction force and torque.

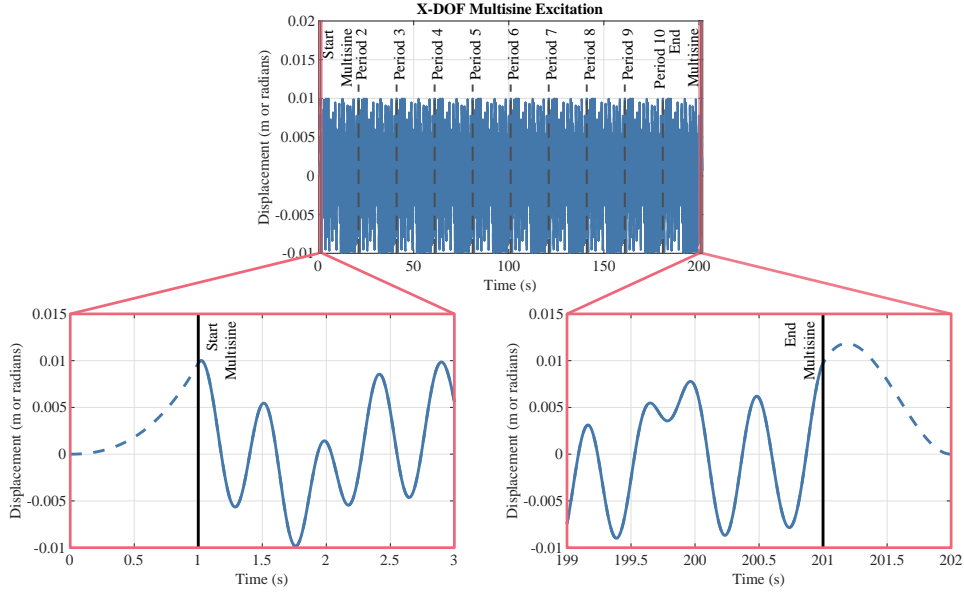

**Figure 8:** Crest factor optimized multisine excitation for SISO system identification. The upper plot shows the ten period multisine signal for the frequency band [0.4, 2.35] Hz at a frequency resolution of 0.05 Hz. The lower plots show a dashed line that represent the ramp in and ramp out part of the signal necessary to transition the hexapod into and out of the multisine; these parts of the signal are excluded from the identification process.

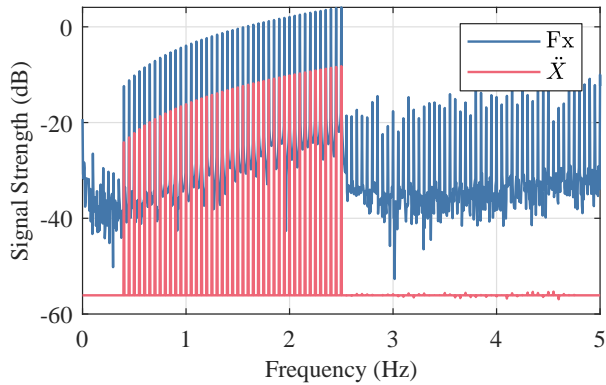

**Figure 9:** SISO FFT of designed input motion and measured output forces.

In figure 9 the fast Fourier transform (FFT) of the input and output data for one of the two experiments is shown. The red curve shows the designed input surge acceleration and the blue curve the unfiltered reaction force in X. The comb-filter-like appearance is a product of transforming multiple periods. When the ten periods are used to compute an averaged FRM, the frequency bin width is equal to the distance between peaks, creating a continuous spectrum. The envelope of the peaks is a designed feature of the phase-optimized multisine. Without optimization, peaks would grow in height with the square of frequency – a behavior that should be avoided to achieve constant uncertainty for the FRF across the frequency spectrum, as explained in the System Identification section. Figure 10 shows the results from the system identification procedure. Figures 10a and 10c show the magnitude and phase angle of the estimated FRF and figures 10b and 10d show the estimated total mass compared to the known mass and the estimated damping. Note that a delay of 41.1 ms between the trigger signal and the beginning of hexapod motion was inferred from the phase relationship between forces and calculated accelerations. The results in this section were calculated with the 41.1 ms phase shift removed. The mass results are very close to the expected result of 16.67 kg with a mean and median measured mass of 16.61 kg and 16.77 kg respectively and root mean square error (RMSE) of 0.882 kg. The linear trend in the FRF phase and the correlation between damping and frequency suggests that a frequency-dependent phase shift still remains, though its source is unclear. Nevertheless, the goal of recovering the mass of the system was successfully met.

After dry SISO multisine tests, water was added to the towing tank and results for the submerged cylinder were collected for SISO multisine excitation signals in surge and sway. The top of the cylinder was submerged 17.5 cm from the surface of the water, so some free-surface interaction was to be expected. The open-source panel code NEMOH was used to estimate the added mass matrix for a truncated cylinder at the same depth beneath the free surface as a comparison (Babarit and Delhommeau, 2015; Kurnia and Ducrozet, 2022; Kur, 2022). Figure 11 shows the results for the SISO prediction of the total mass and damping of the submerged cylinder. Although the consistency between the results for surge and sway motion matches very closely for both the total mass and total damping, they significantly under-predict the added mass and over-predict the added damping compared to the panel code results. An exhaustive debugging process was undertaken that eventually concluded that one of the strain gauges in the load cell was damaged or otherwise producing erroneous signals.

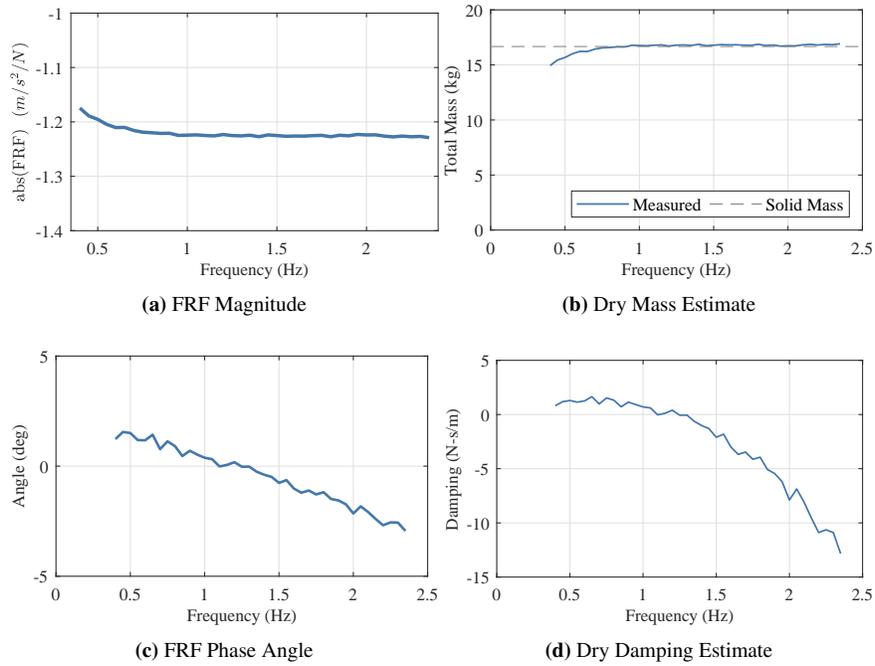

**Figure 10:** Example of a dry cylinder SISO multisine experiment using the hexapod. Figures (a) and (c) show the magnitude and phase angle of the estimated FRF. Figures (b) and (d) show the estimated dry mass of the cylinder and damping. The SISO multisine is able to correctly recover the mass of the dry cylinder; however, the damping results may show that some unaccounted for delay may still be present in the system.

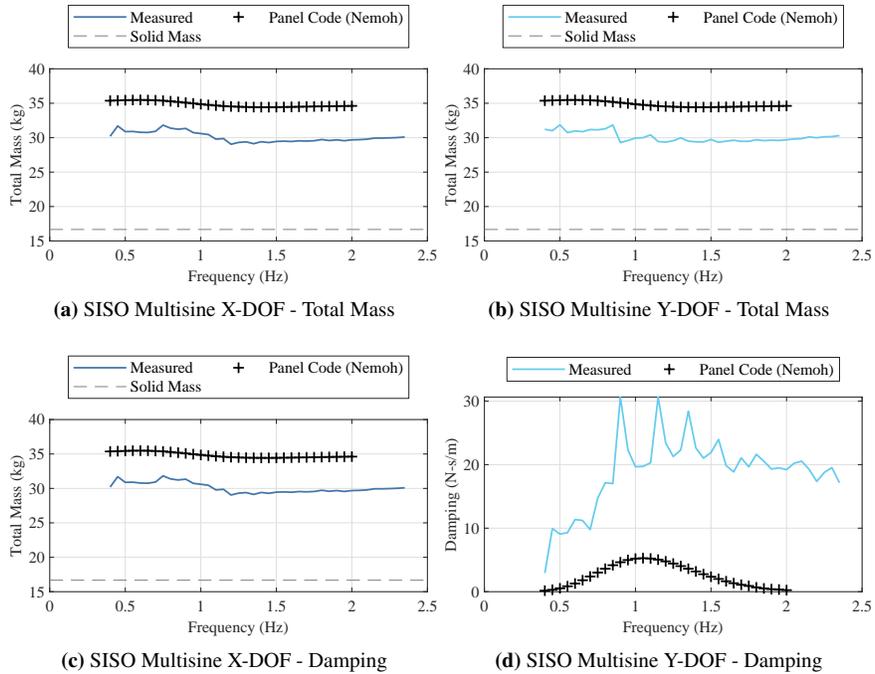

**Figure 11:** SISO predictions for added mass and damping of the submerged cylinder system. Figures (a) and (c) are the results for motion on the X-axis, and figures (b) and (d) are the results for motion on the Y-axis. The measured values are represented by the solid line, the +'s represent the value predicted by the Nemoh panel code, and for the two mass figures, the dashed line shows the solid mass.

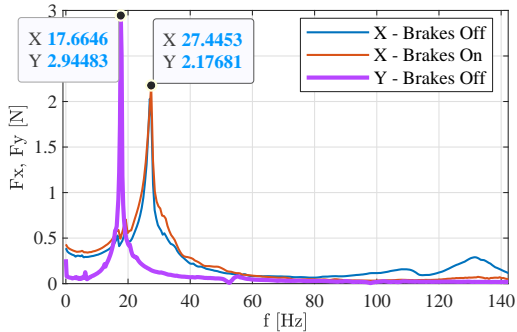

**Figure 12:** Hammer impact test on the cylinder system. The findings indicate that neither the strut nor the PID control compliance affects the system's response within the attainable frequency range.

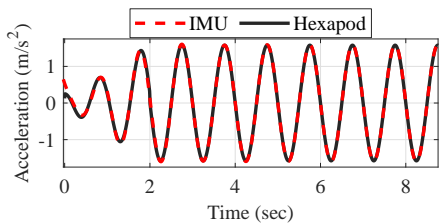

**Figure 13:** Comparison of measured (IMU) and designed (Hexapod) accelerations in X during sinusoidal surge motion. The signal amplitudes differ by only 0.7%

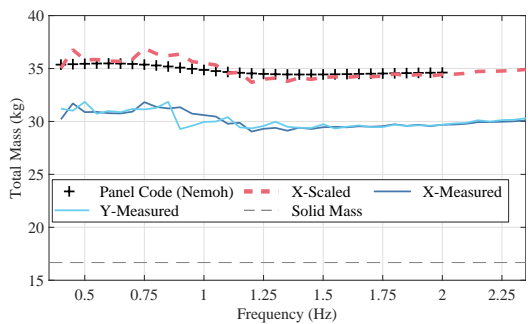

**Figure 14:** Total mass of cylinder system with a scaling of 115% applied. Scaled result matches the panel code result quite well qualitatively.

**Debugging Measures**

The hypothesized contributors to the estimation error were: too much compliance in the "sprung" mass due to a flexible strut, compliance in the PID controller, inaccurate platform motions, incorrect accelerations at the center of gravity, or issues with the load cell itself. Each possible cause was checked in turn. To measure the compliance of the strut and PID controller, hammer-impact tests were undertaken with both the actuator brakes engaged – to measure strut compliance – and disengaged – to measure PID compliance. The results of the impact tests depicted in figure 12 show that both the strut and the PID controller are very stiff, with resonances one to two orders of magnitude above the bandwidth of the hexapod motions.

The kinematics of the hexapod were evaluated next. A comparison was made between the commanded motions and camera tracking results for surge amplitudes between 5mm and 4 cm. The observed RMSE between the commanded amplitude and the observed amplitude was 0.385 mm, and therefore this was also deemed an unlikely source of the error. To verify accelerations, an Advanced Navigation Orientus IMU was affixed directly to the cylinder. Figure 13 shows the expected and measured surge accelerations, showing an error of less than 0.7%.

With compliance and kinematics eliminated as sources of error, the last remaining source of error was the load cell itself. While it was observed that the behavior of the measurements in the surge and sway motion remained linear, measurements conducted with the cylinder empty and full of water showed that positive loading in the $z$-axis produced a decrease in the sensitivities for forces in $x$ and $y$, indicating a damaged strain gauge. If the submerged measurement results are scaled by a factor of 115% as shown in figure 14 it is seen that the SISO submerged surge tests appear to match the panel code quite very closely.

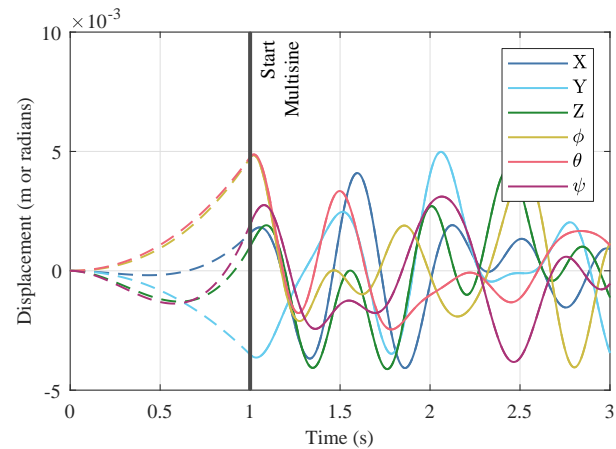

**Figure 15:** Example of one MIMO experiment phase-optimized multisine excitation. Frequency band [0.4, 2.35] Hz at a frequency resolution of 0.05 Hz.

**Qualitative MIMO Multisine Results**

Due to the load cell malfunction, quantitative testing is impossible until the necessary repairs and recalibration are completed. However, it was shown that the qualitative characteristics of the added mass results from the system identification method do appear to be correct and so tests of the MIMO multisine excitation and system identification were performed for both the dry condition and submerged condition. An example of one of the MIMO multisine excitation signals is shown in figure 15 where the six displacement inputs at the beginning of the excitation signal are shown to diverge mutually orthogonal multisines.

Figure 16 shows the full mass matrix with respect to frequency for the cylinder suspended in air. Observe that the surge and sway values match those of the SISO multisine experiments almost exactly. Furthermore, the moments of inertia are well captured, although some deviation at lower frequencies is present. This is likely the result of very low rotation amplitudes injecting sub-optimal power into the lower frequency bands, relative to the translational motions, which can be corrected by changing the scale value used in generating the rotation amplitudes in the HMG application. Coupling terms $M_{15}$ and $M_{24}$ both demonstrate errors that, while large, are consistent with the hypothesized attenuation of horizontal forces by the load cell.

Figures 17 and 18 show the total mass and added damping matrices for the submerged cylinder system between [0.4, 2.35] Hz. Qualitatively, the total mass trends seen from the SISO multisine experiments continue here, following the trends – if not the values – of the panel code for the translational DOFs. Most of the off-diagonal elements are also correctly resolved as zero. For the rotational DOFs, the values are considerably over-predicted on the diagonals and have a strong frequency dependence. Because the measured forces and torques are translated to the motion datum, torque measurements are negatively affected by the faulty load cell signals. The damping estimates need improvement; there is a clear trend of phase delay along the diagonal, which could be resolved by using measured accelerations instead of the prescribed kinematics to eliminate the influence of the PID controller from the system identification procedure.

**CONCLUSIONS & DISCUSSION**

A novel six-DOF motion platform known as a hexapod was designed, fabricated, programmed, and installed at the IIHR towing tank at The University of Iowa. The hexapod is capable of carrying models in excess of 500 kg and executing large amplitude motions of 25 - 35 cm for the linear DOFs and 20° - 45° for the angular DOFs with high precision and repeatability. The hexapod is capable of a frequency range from 0.25 to 5 Hz and a maximum velocity at the platform center of 1 m/s. The hexapod adds significant capabilities to the IIHR towing tank including static positioning for models and sensors, non-planar motions of surface and subsurface vessels, motion emulation, as well as rapid system identification. The system identification developed for the hexapod utilizes phase-optimized multisines that – when multiple DOFs are considered – are mutually orthogonal in time and frequency domains, allowing for simultaneous excitation of all DOFs. A truncated vertical cylinder in dry and submerged conditions was used as a test geometry. Unfortunately, following an exhaustive debugging of the hexapod system and data processing, a defective strain gauge in the load cell was determined the likely cause of under-estimated horizontal forces by up to 30%. The result was an underestimationion of the added mass and an overestimationion of the added damping for the cylinder when compared with well-established panel code predictions. Quantitative errors notwithstanding, the trends observed in the mass matrix largely match those of the panel code. This is an encouraging result, especially considering that all the full 6×6 mass and damping matrices were resolved across a range of frequencies in less than a cumulative hour of data collection.

While tests reported in this work were limited to a truncated cylinder, the platform should be applicable to a wide range of model geometries and sizes. Despite the relatively modest cross section of the IIHR towing tank, prior campaigns have tested models up to 3.048 m (10 ft) in length (Yoon *et al.*, 2015), which may be taken as a practical upper limit on model length. The hexapod is currently tuned to handle models with a mass of approximately 100 kg, but is capable of dynamic motions with a payload of up to 500 kg. For submersible models especially, the model sizes will likely be limited by the blockage ratio.

When addressing how many individual experiments are necessary for the identification and the length of data collection required for both sufficient identification and accuracy depend on three criteria: the lowest frequency to be resolved, the desired frequency resolution, and the number of DOFs. For a 6-DOF MIMO experiment, if the lowest frequency is 0.2 Hz, one period of the excitation signal is 5 seconds. Simulation testing has indicated that the robust local polynomial method requires at least 10 period and 2 realizations, resulting in data collection of 200 seconds per experiment for each DOF. A detailed sensitivity analysis with the validation cylinder test case should be undertaken. Statistical criteria have not been fully designed yet, though these will be pursued after the cylinder validation is completed. These criteria must be established to strike a balance between testing time, frequency resolution, and robust noise-rejection. The estimation of noise covariances by the robust polynomial FRM estimator is a useful feature that will allow signals to be designed to achieve desired signal-to-noise ratios and FRM uncertainty bounds. When testing with forward speed, the test matrix would be repeated at each speed to produce a locally linearized model. For cases with forward speed, it is expected that the added mass should remain approximately constant, allowing the added stiffness and added mass to be quantified to be easily found. Memory effects associated with forward speed, such as the production of vortical wakes, should be captured as frequency-dependent damping terms, provided the wake dynamics are approximated as linear and wake history effects remain periodic. In the case of non-periodic forces produced by turbulent flow, the robust local polynomial FRM estimation includes averaging to remove the effects of random noise. Finally, testing at forward speed will limit the available time for data collection; to accommodate this limitation, the multisine parameters may be optimized to fit one realization of an experiment into the time available at steady forward speed, and both experiments and realizations may be distributed across several runs.

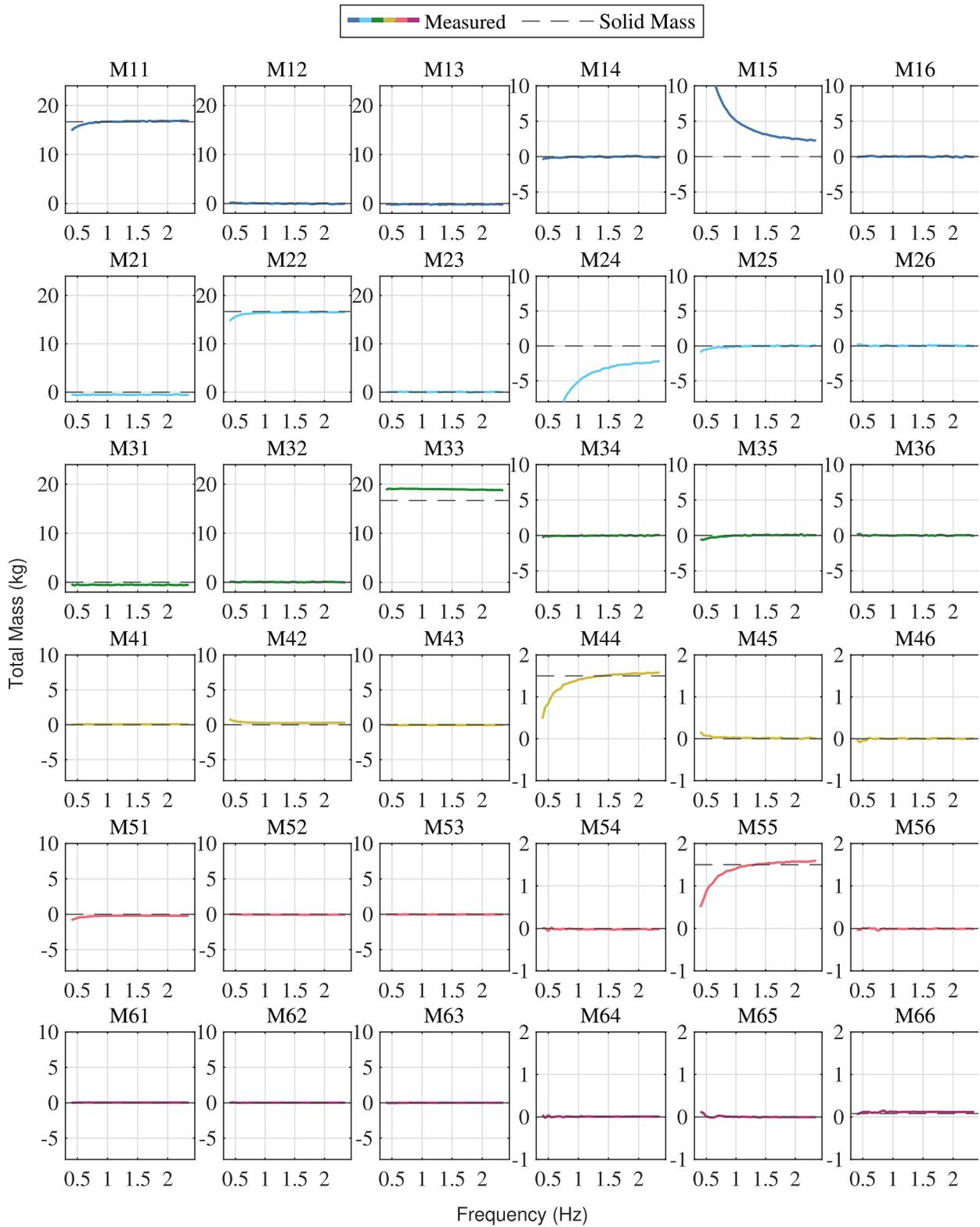

**Figure 16:** MIMO measured mass for vertical cylinder system hanging in air. Dashed lines shows the true values for the diagonal elements, which are the only nonzero elements for motions about the center of gravity.

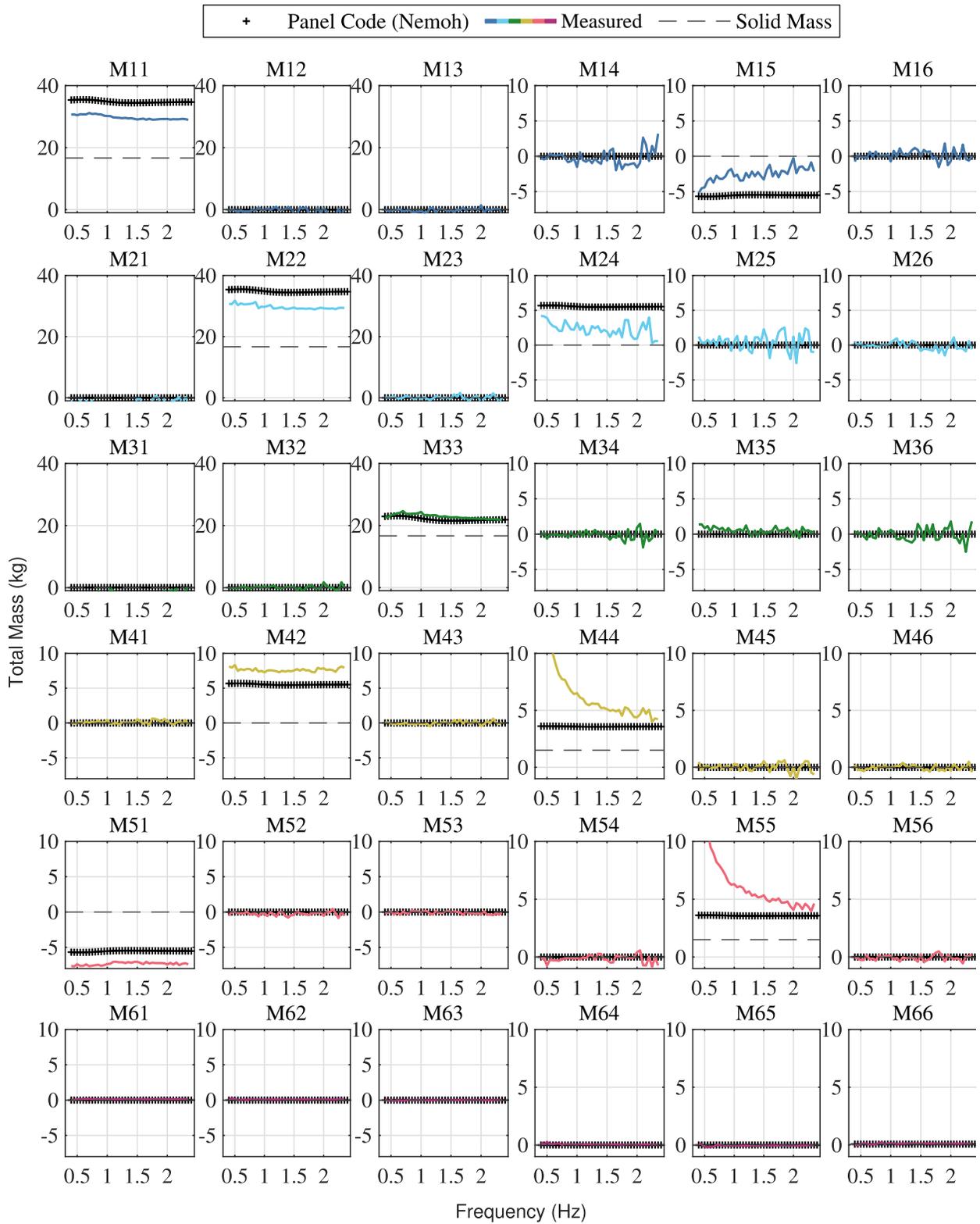

**Figure 17:** MIMO measured mass for submerged cylinder system in water. The +'s show values predicted by the Nemoh panel code.

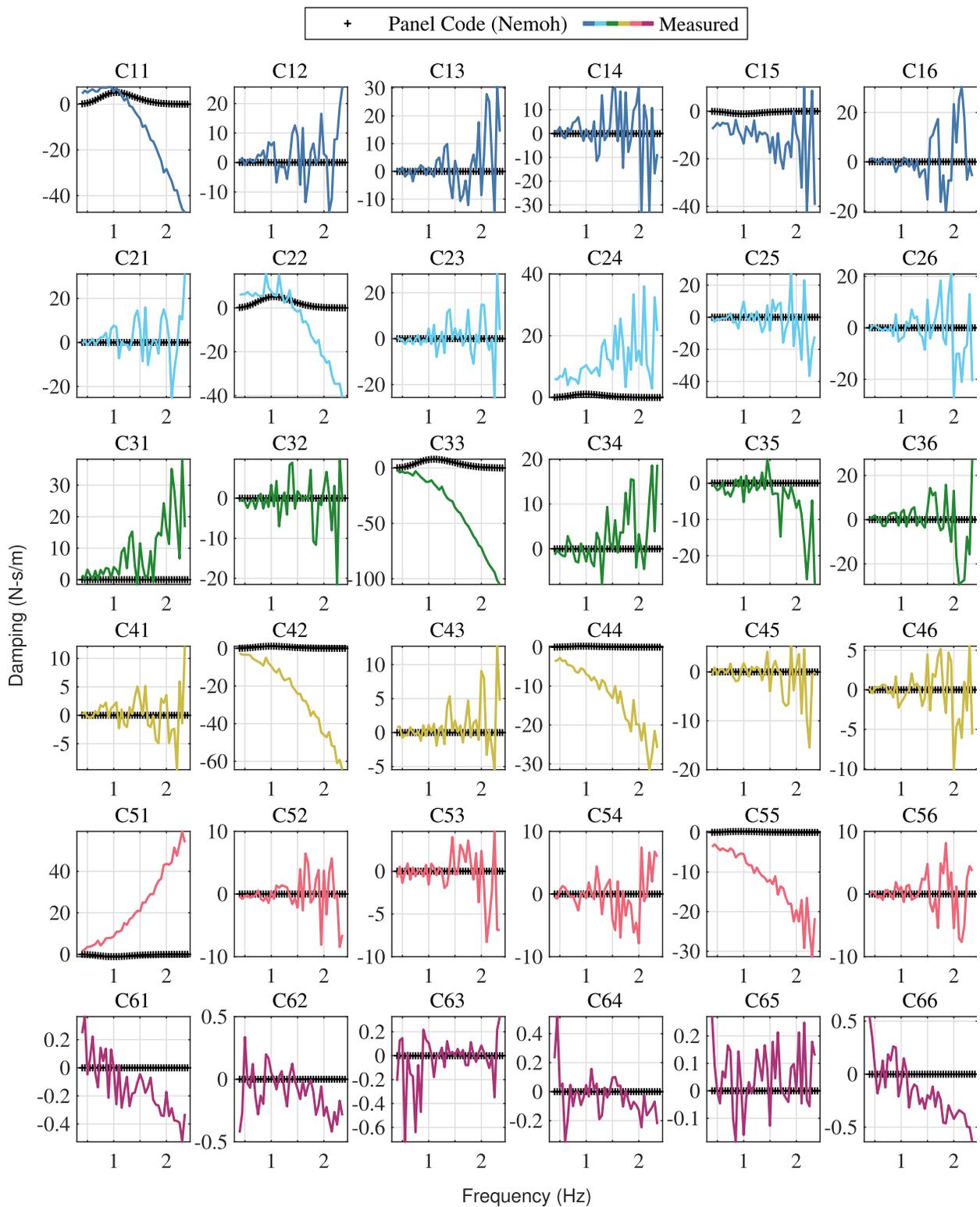

**Figure 18:** MIMO measured damping for submerged cylinder system in water. The +'s show values predicted by the Nemoh panel code.

The defective load cell significantly impeded progress for several weeks, with the complexity of the hexapod system and its identification methodology complicating bug-isolation, as detailed in Debugging Measures. Furthermore, there still likely exist– albeit minor – bugs in the system, either due to phase lag in the PID controller, phase delay in long cable runs, or asynchronous execution of threads on the motion controller. Despite these challenges, measurements are planned with a repaired load cell and a permanent platform-mounted IMU. Using IMU measurements as the system inputs in lieu of the prescribed accelerations should signficantly enhance the precision of the estimated added mass and added damping. Following the re-verification with the submerged cylinder, tests are planned to develop a system model for maneuvering UUVs. If successful, this represents an empirical approach to maneuvering model development that competes with panel codes in terms of speed, but which can be used for more complicated geometries or in viscosity-dominated flows.

## Acknowledgments

This work is supported by the U.S.office of Naval Research (ONR) through grant N00014-22-1-2097, managed by Troy Hendricks.